\documentclass[12pt]{article}
\usepackage{amsmath}
\usepackage{graphicx,psfrag,epsf,bm, soul,color,tablefootnote}
\usepackage{enumerate}
\usepackage{natbib}
\usepackage{url} % not crucial - just used below for the URL 
\usepackage{graphicx}
\usepackage{comment}
\usepackage{hyperref}
\hypersetup{
    colorlinks=true,
    linkcolor=black,
    filecolor=black,      
    urlcolor=blue,
    citecolor=black,
    pdftitle={Overleaf Example},
    pdfpagemode=FullScreen,
    }

\usepackage{appendix}
\usepackage{hyphenat}
\usepackage{enumitem}
\setlist{leftmargin=20mm}
\usepackage[margin = 20mm]{geometry}
\graphicspath{ {./figures/} }
\usepackage[singlelinecheck=false % <-- important
]{caption}
\usepackage{setspace}
\usepackage{authblk}
\newcommand{\nonl}{\renewcommand{\nl}{\let\nl\oldnl}}% Remove line number for one line
\begin{document}

% \journaltitle{Journal Title Here}
% \DOI{DOI HERE}
% \copyrightyear{2022}
% \pubyear{2019}
% \access{Advance Access Publication Date: Day Month Year}
% \appnotes{Paper}

% \firstpage{1}

%\subtitle{Subject Section}

% \title{\bf A Matrix Ensemble Kalman Filter-based Multi-arm  Neural Network to Adequately Approximate Deep Neural Networks}%{In-Situ Stacking of Deep Neural Networks using Ensemble Kalman Filters}
\def\spacingset#1{\renewcommand{\baselinestretch}%
{#1}\small\normalsize} \spacingset{1}

% \doublespacing
% \singlespacing

%%%%%%%%%%%%%%%%%%%%%%%%%%%%%%%%%%%%%%%%%%%%%%%%%%%%%%%%%%%%%%%%%%%%%%%%%%%%%%

\title{\bf A Matrix Ensemble Kalman Filter-based Multi-arm  Neural Network to Adequately Approximate Deep Neural Networks}
\author[1]{Ved Piyush}
\author[2]{Yuchen Yan}
\author[3]{Yuzhen Zhou}
\author[2]{Yanbin Yin}
\author[1]{Souparno Ghosh}
\affil[1]{Department of Statistics, University of Nebraska, Lincoln}
\affil[2]{Nebraska Food for Health Center, Department of Food Science and Technology, University of Nebraska, Lincoln}
\affil[3]{Amazon, Bellevue, Washington}
\date{}
\maketitle

% \author[1,$\ast$]{Ved Piyush}
% \author[2]{Yuchen Yan}
% \author[1]{Yuzhen Zhou}
% \author[2]{Yanbin Yin}
% \author[1]{Souparno Ghosh}

% \authormark{Piyush et al.}

% \address[1]{\orgdiv{Department of Statistics}, \orgname{University of Nebraska, Lincoln}, \orgaddress{\street{340 Hardin Hall North Wing}, \postcode{68583}, \state{Nebraska}, \country{USA}}}
% \address[2]{\orgdiv{Nebraska Food for Health Center, Department of Food Science and Technology}, \orgname{University of Nebraska, Lincoln}, \orgaddress{\street{1901 N 21 ST}, \postcode{68588}, \state{Nebraska}, \country{USA}}}

% \corresp[$\ast$]{Corresponding author: \href{email:ved@huskers.unl.edu}{ved@huskers.unl.edu}}

% \received{Date}{0}{Year}
% \revised{Date}{0}{Year}
% \accepted{Date}{0}{Year}

%\editor{Associate Editor: Name}

%\abstract{
%\textbf{Motivation:} .\\
%\textbf{Results:} .\\
%\textbf{Availability:} .\\
%\textbf{Contact:} \href{name@email.com}{name@email.com}\\
%\textbf{Supplementary information:} Supplementary data are available at \textit{Journal Name}
%online.}

\begin{abstract}
\singlespacing
\normalsize
Deep Learners (DLs) are the state-of-art predictive mechanism with applications in many fields requiring complex high dimensional data processing. Although conventional DLs get trained via gradient descent with back-propagation, Kalman Filter (KF)-based techniques that do not need gradient computation have been developed to
approximate DLs. We propose a multi-arm extension of a KF-based DL approximator that can mimic DL when the sample
size is too small to train a multi-arm DL. The proposed Matrix Ensemble Kalman Filter-based multi-arm ANN (MEnKF-
ANN) also performs explicit model stacking that becomes relevant when the training sample has an unequal-size feature set. Our proposed technique can approximate Long Short-term Memory (LSTM) Networks
and attach uncertainty to the predictions obtained from these LSTMs with desirable coverage. We demonstrate how MEnKF-ANN can “adequately” approximate an LSTM network trained to classify
what carbohydrate substrates are digested and utilized by a microbiome sample whose genomic sequences consist of
polysaccharide utilization loci (PULs) and their encoded genes. The scripts to reproduce the results in this paper are available at \href{https://github.com/Ved-Piyush/MEnKF-ANN-PUL} {https://github.com/Ved-Piyush/MEnKF-ANN-PUL}. \\
\end{abstract}
% \clearpage
\maketitle

\section{Introduction}\label{sec:intro}
Deep Learners (DLs) have achieved state-of-art status in empirical predictive modeling in a wide array of fields. The ability of DLs to synthesize vast amounts of complex data, ranging from high dimensional vectors to functions and images, to produce accurate predictions has made them the go-to models in several areas where predictive accuracy is of paramount interest. Bioinformatics has also seen a steep increase in articles developing or deploying DL techniques in recent years \citep{min2017deep}.
However, conventional DLs trained via gradient descent with back-propagation require tuning of a large number of hyperparameters. Additionally, given the vast number of weights that DLs estimate, they are prone to overfitting when the training sample size is relatively small. Since the gradient descent algorithms compute the weights deterministically, DLs in their vanilla form do not yield any uncertainty estimate.

Several techniques have been proposed to alleviate the foregoing issues in DLs. For instance, Bayesian Neural Network (BNN) \citep{kononenko1989bayesian, neal2012bayesian} was explicitly devised to incorporate epistemic and aleatoric uncertainty in the parameter estimation process by assigning suitable priors on the weights. The Bayesian mechanism can process these priors and generate uncertainty associated with DL predictions. Additionally, with judicious choice of priors, BNNs can be made less prone to overfitting \citep{fortuin2021bayesian, srivastava2014dropout}. Variational inference is another popular technique for uncertainty quantification in DLs. In particular, \cite{hinton1993keeping} showed that a posterior distribution for the model weights could be obtained by minimizing the Kullback-Leibler distance between a variational approximation of the posterior and the true posterior of the weights. The Bayes by Backprop \citep{blundell2015weight} is another technique that uses variational formulation to extract uncertainty associated with the weights in DLs.

However, the Monte Carlo dropout technique \citep{srivastava2014dropout}, wherein each neuron (and all of its connections) is randomly dropped with some probability during the model training process, has arguably turned out to be the most popular method to regularize DLs and extract predictive uncertainty. In addition to the conceptual simplicity of the dropout technique, it was shown that the models trained using dropout are an approximation to Gaussian processes and are theoretically equivalent to variational inference \citep{gal2016dropout}. %By applying the Monte Carlo Dropout technique during the prediction process (after the model has been trained), many different thinned copies of the trained neural network can be obtained. Thus using these many different thinned copies, a distribution of predictions can be obtained for the same testing sample. \cite{hinton1993keeping} developed a method based on variational inference. They showed that a posterior distribution for the model weights could be obtained by minimizing the Kullback-Leibler distance between a variational approximation of the posterior and the true posterior of the weights. 
Regardless of its conceptual simplicity and theoretical underpinning, dropout methods require gradient computation 
% over many different thinned copies of the trained DLs to obtain the posterior distribution of weights and predictive distribution of the test samples
. Hence, it can be quite computationally intensive in DLs with millions of parameters.
%This reliance on gradient calculations can also prove challenging in using these two methods to train deep neural networks and Recurrent Neural Networks, where vanishing and exploding gradients are common phenomena \citep{hochreiter1998vanishing}. 

Another suite of methods for approximating DLs uses Kalman Filters (KF), or its variants, to obtain approximate estimates of the DL parameters \citep{yegenoglu2020ensemble, rivals1998recursive, wan2000unscented, julier2004unscented, chen2019approximate}. In particular, the Ensemble Kalman Filter (EnKF) technique offers a computationally fast approximation technique to DLs.   For instance, \cite{chen2019approximate} train a single hidden layer neural network using the EnKF updating equations outlined in \cite{iglesias2013ensemble} and show how using the augmented state variable, one can estimate the measurement error variance.  In the DL setting, \cite{chen2018approximate} demonstrate the utility of EnKF in approximating a Long Short Term Memory (LSTM) model. \cite{yegenoglu2020ensemble} use the EnKF to train a Convolutional Neural Network directly using the Kalman Filtering equations derived in \cite{iglesias2013ensemble}.  All these methods approximate single-arm DLs and, therefore, cannot be used in situations where input features can be represented in multiple ways. Our target is to develop a multi-arm approximator to the DL. In principle, we can train different DL approximators for each representation and perform a post-training model averaging. However, that would increase the computation cost substantially. We argue that since multiple different feature representations do not necessarily offer complimentary information, developing a multi-arm approximator that performs model averaging while training would perform ``adequately''. To that end, we develop a Matrix Ensemble Kalman Filter (MEnKF)-based multi-arm ANN that approximates a deep learner and simultaneously performs model averaging. 

We apply our method to approximate an LSTM model trained to classify what carbohydrate substrates are digested and
utilized by a microbiome sample characterized by genomic sequences consisting of polysaccharide utilization loci (PULs)  \citep{bjursell2006functional}
and their encoded genes. We use two different representations of the genomic sequences consisting of the PULs in two different arms, our MEnKF-ANN approximator, and demonstrate that our approximator closely follows the predicted probabilities obtained from the trained LSTM. We also generate prediction intervals around the LSTM-predicted probabilities. Our results show that the average width of the prediction interval obtained from the MEnKF-ANN approximator is lower than that obtained from the original LSTM trained with MC dropout.
% but the computation time for the former is considerably less as compared to that of the latter.
We also perform extensive simulations, mimicking the focal dataset, to demonstrate that our method has desirable coverage for test samples compared to the MC dropout technique. 
%Once again, the average width of the prediction intervals obtained from these two techniques is comparable.
Finally, we emphasize that even though the original problem is binary classification, our MEnKF-ANN approximator is designed to emulate the probabilities obtained from the original LSTM model and quantify the uncertainties in the LSTM-predicted probabilities.  

The remainder of the article is organized as follows: In section \ref{sec:datadesc}, we describe the aspects of an experimentally obtained microbiome dataset that motivated us to design this approximator. In section \ref{sec:background}, for the sake of completion, we offer a brief review of KF, EnKF, and how these techniques have been used to train DLs. Section \ref{sec:method} details the construction of our MEnKF-ANN method. In section \ref{sec:simulation}, we offer extensive simulation results under different scenarios and follow it up with the application on real data in section \ref{sec:application}. Finally, section \ref{sec:discussion} offers concluding remarks and future research directions.

\section{Motivating Problem}\label{sec:datadesc}
The human gut, especially the colon, is a carbohydrate-rich environment \citep{kaoutari2013abundance}. However, most of the non-starch polysaccharides (for example, xylan, pectin, resistant glycans) reach the colon undegraded \citep{pudlo2022phenotypic} because human digestive system does not produce the enzymes required to degrade these polysaccharides \citep{flint2012microbial}. Instead, humans have developed a complex symbiotic relationship with gut microbiota, with the latter providing a large set of enzymes for degrading the aforementioned non-digestible dietary components \citep{valdes2018role}. %But, to be effective, these microbes need to adhere to a substrate (or a few substrates) and tolerate the conditions generated during the degradation process \citep{koropatkin2012glycan}.  
Consequently, an essential task in studying the human gut microbiome is to predict what carbohydrate substrates a microbiome sample can digest from the genetic characterization of the said microbiome \citep{koropatkin2012glycan}. 

In order to generate a focused genetic characterization of the microbes that relates to their carbohydrate utilization property, one often investigates the genes encoding the Carbohydrate Active Enzymes (CAZymes) and other proteins that target glycosidic linkages and act to degrade, synthesize, or modify carbohydrates \citep{lombard2014carbohydrate,zhang2018dbcan2}. This set of genes tend to form physically linked gene clusters in the genome known as polysaccharide utilization loci (PULs) \citep{bjursell2006functional}. Consequently, the gene sequences associated with PULs of microbes could be used as a predictor to ascertain the carbohydrate substrate the microbe can efficiently degrade. However, these gene sequences are string-valued quantities \citep{huang2018dbcan, stewart2018open} and hence their naive quantitative representations (for instance, one-hot-encoding or count vectorization)  often do not produce classifiers with acceptable accuracy \citep{badjatiya2017deep}. Instead, we can use LSTM to process the entire sequence of string-valued features and then implement a classifier with a categorical loss function. The trained LSTM, by default, produces an embedding of the gene sequences in a vector space. Alternatively, we can also use a Doc2Vec embedding of the entire sequence associated with the PUL or an arithmetic average of Word2Vec embedding of each gene in the sequence and train a shallow learner (an ANN, for example). Since various representations of the predictors are available,  we can train a multi-arm DL that takes different representations of features in different arms and performs concatenation/late integration of the embeddings before the prediction layer \citep{liu2020deepcdr, sharifi2019moli}. However, such multi-arm DLs require a relatively large number of training samples - typically tens of thousands. 

Since the experimental characterization of new PULs for carbohydrate utilization is an expensive process \citep{ausland2021dbcan}, we do not have large enough labeled samples to train complex multi-arm DLs. This predicament motivates us to develop a multi-arm approximator to a DL with the following capabilities: (a) it must ``adequately'' approximate the focal single-arm DL, (b) it should be able to ingest different feature representations in different arms and perform model averaging, (c) it should be able to detect if the set of representations supplied to it is substantially different from the representations used to train the original DL, i.e., sensitive to major misspecification. Since the original DL is trained on a single representation of features and the approximator is ingesting multiple representations, the latter is misspecified in a strict sense. However, this misspecification is deliberately introduced to circumvent training a multi-arm DL and assess, via model averaging, whether there is any benefit in using multiple representations of the same feature set. 

\begin{figure}[!t]%
\centering
\includegraphics[width = 0.7\textwidth]{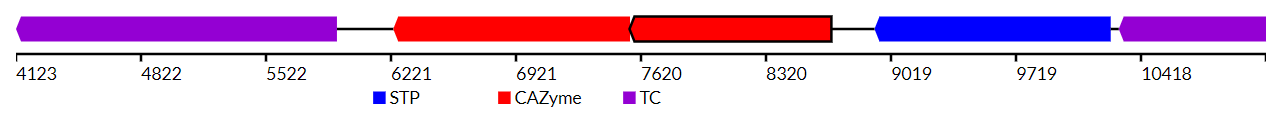}
\caption{Pectin PUL}%\label{fig1}
\label{fig:pul}
\end{figure}

\begin{figure}[!t]%
\centering
\includegraphics[width =  0.7\textwidth]{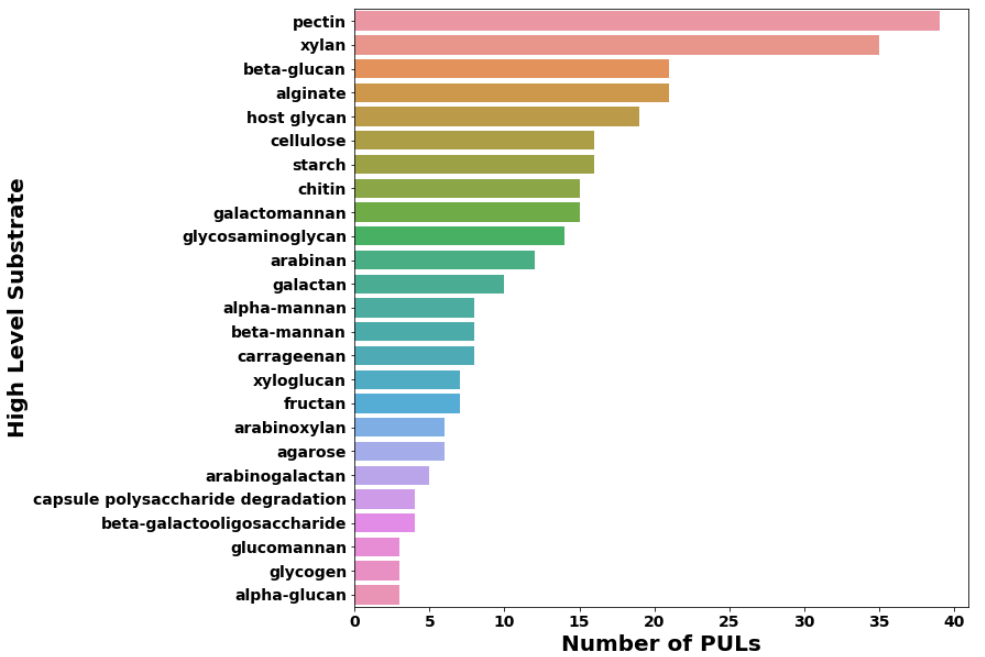}
\caption{Frequency distribution for the various substrates}%\label{fig1}
\label{fig:freqdist}
\end{figure}

\begin{figure}[!t]%
\centering
\includegraphics[width =  0.7\textwidth]{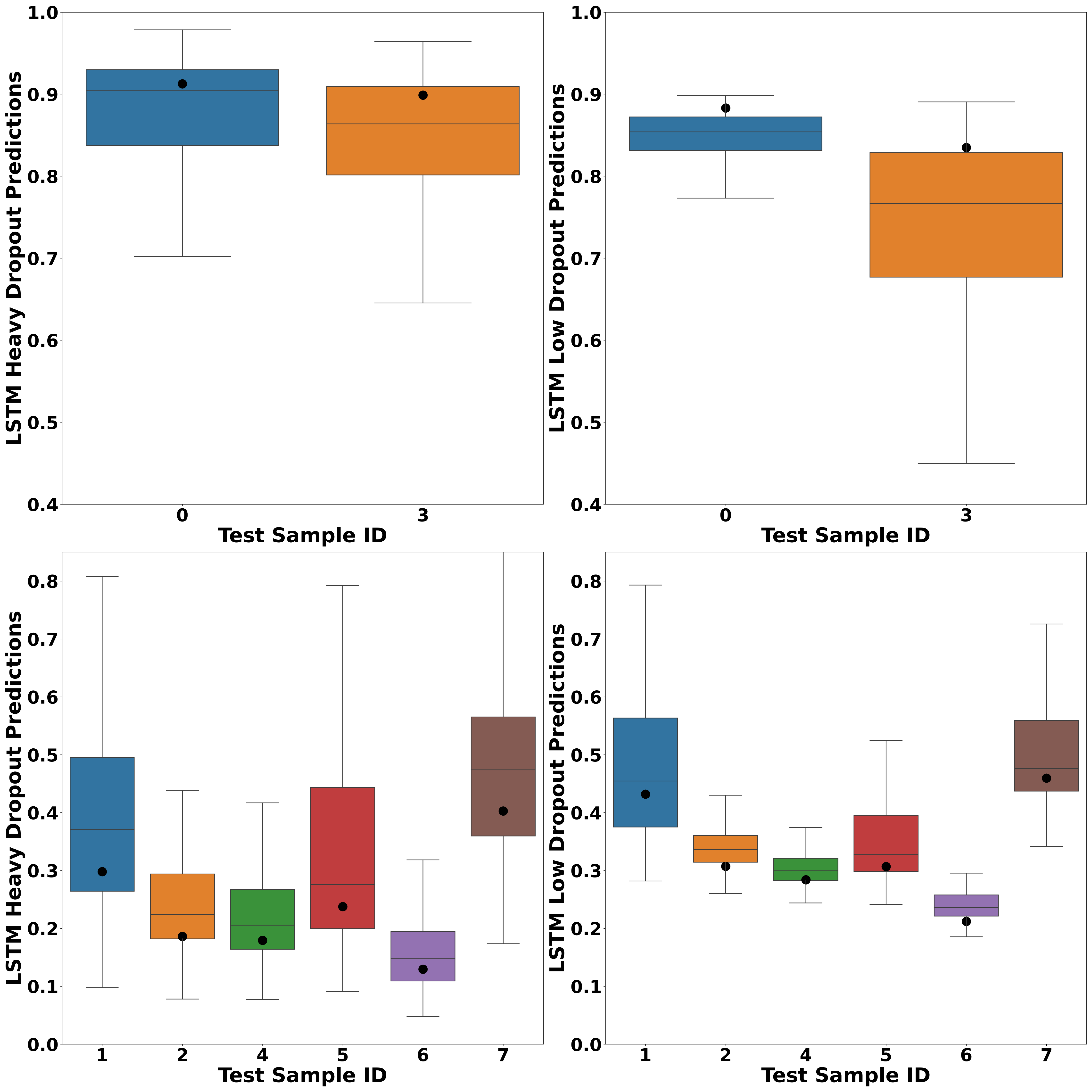}
\caption{Boxplots showing the Predictions superimposed with the ground truth value from Heavy and Low Just LSTMs}%\label{fig1}
\label{fig:boxplots2}
\end{figure}

We extract the dataset from the dbCAN-PUL database \citep{ausland2021dbcan} that contains experimentally verified PULs and the corresponding GenBank sequences of these PULs along with known target carbohydrate substrates. Figure \ref{fig:pul} shows an example of a gene sequence associated with a PUL for the substrate Pectin. %As mentioned earlier, data on such experimentally derived PUL along with their target  substrate is  limited, 
We have a total of approximately 411 data points. Figure \ref{fig:freqdist}  shows the dataset's frequency distribution of various target substrates. We do not have sufficient samples to train a complex DL to classify all the available substrates. Hence we propose to classify the two most frequently occurring target substrates - Xylan and Pectin - and train an  LSTM binary classifier. Seventy-four samples belong to these two classes of substrates in a reasonably balanced way. One way to attach uncertainty to the probabilities estimated by the LSTM architecture is to activate the dropout layers during the prediction phase. This will generate multiple copies of the prediction depending on the architecture of the dropout layers. However, we need to decide on how many dropout layers to include and where to place them. Often the prediction intervals are pretty sensitive to the number and placements of dropout layer(s). For instance, the top left and bottom left panels of Figure \ref{fig:boxplots2} show the prediction intervals associated with eight held-out test samples obtained when two dropout layers were included  - one inside the LSTM and one just before the final prediction layer.
In contrast, the top right and bottom right panels of Figure \ref{fig:boxplots2} show the prediction intervals associated with the same test samples obtained when the second dropout layer was removed from the foregoing LSTM architecture. Observe how the number and placement of dropout layers influence the variability in the width of these intervals. If we wish to control the width variability, the placement of the dropout layer becomes a tuning parameter and further increases the hyperparameter search space. %extensive tuning is required to answer these questions and 

We empirically show that our MEnKF-ANN approximator, trained on the logit transformed LSTM-estimated probabilities as the response and the embedding of the sequences obtained from LSTM and Doc2Vec operations as two types of features produce more stable prediction intervals regardless of the location of the dropout layer in the original LSTM. %Although, our procedure requires an additional round of training, but its ability to adeqbut we show this procedure the approximator has only two key hyperparamaters- the size of the ensemble and the initial variance of the ensemble members. Consequently, a simple grid search could be used to tune these hyperparameters in terms of prediction adequacy. %Thus, from practical perspective our approximator not only alleviates the necessity to train multi-arm DL but also offer substantial computational benefit.

\section{Background}\label{sec:background}
This section offers a brief overview of Kalman Filters and Ensemble Kalman Filters and discusses how these methods have been used to train NNs and approximate DLs. For an extensive discussion on KF and EnKF technique, we direct the audience to \cite{katzfuss2016understanding} and  \cite{evensen2003ensemble}, respectively.
\subsection{Linear State Space Model \& Kalman Filter}\label{sec:kf}
Consider a linear Gaussian state-space model given by

\begin{align}
\label{eq:1}
y_t = H_{t}x_t + \epsilon_t, \ \ \epsilon_t \sim \mathcal{N}_{m_{t}}(0, R_t)\\
\label{eq:2}
x_t = M_tx_{t-1} + \eta_t, \ \ \eta_t \sim \mathcal{N}_{n}(0, Q_t)
\end{align}

\noindent
where $y_t$ is the $m_t$ dimensional observation vector at time step $t$, $x_t$ is the $n$ dimensional state variable at that time,  $H_t$ and $M_t$ denote the observation and the state transition matrices. %$H_t$ describes how the state variable $x_t$ relates to the observation and $M_t$ denotes how the state vector at time $t-1$ is related to the state vector at time $t$. $v_t$ and $w_t$ are error sequences drawn from a multivariate normal distribution with zero mean vector and covariance matrix $R_t$ and $Q_t$, respectively. It is assumed that the matrices $M_t$, $H_t$, $R_t$, and $Q_t$ are known.  See \cite{katzfuss2016understanding} for a review. 

%An important problem in the state-space model is ascertaining the state variable's filtering distribution at time $t$ conditional on the measurements till time $t$. That is to find the conditional distribution of $x_t$ given $y_{1{:}t} := \{y_{1}, y_{2}, ..., y_{t}\}$ represented as $x_{t}|y_{1{:}t}$. This conditional distribution can be calculated using the forecast and update steps of the Kalman Filter. Assuming that the filtering distribution for the state at time step $t-1$ is given by 
\noindent
Assume that the filtering distribution of the state vector at $t-1$ is given by 
\begin{equation}\label{eq:3}
x_{t-1}|y_{1{:}t-1} \sim \mathcal{N}(\hat{\mu}_{t-1}, \hat{\Sigma}_{t-1}),
\end{equation}

\noindent
KF computes the forecast distribution at $t$ using \eqref{eq:2} as 

\begin{align}\label{eq:4}
x_{t}|y_{1{:}t-1} \sim \mathcal{N}(\tilde{\mu}_{t}, \tilde{\Sigma}_{t}) \nonumber\\
\tilde{\mu}_{t} := M_{t}\hat{\mu}_{t-1}, \nonumber\\
\tilde{\Sigma}_{t} := M_{t}\hat{\Sigma}_{t-1}M_{t}^{'} + Q_t
\end{align}

\noindent
Once the measurement at time step $t$ becomes available, the joint distribution of $(x_t,y_t)$ is given by%the Kalman Filter update step modifies the forecast distribution. The joint distribution of $x_t$ and $y_t$ conditional on measurement data till time step $t-1$ i.e $y_{1{:}t-1}$ is derived using the multivariate normal distribution  
\begin{equation}\label{eq:5}
\begin{pmatrix}
x_t \\
y_t
\end{pmatrix} \bigg| y_{1{:}t-1}  \sim N\left(\begin{pmatrix}
\tilde{\mu}_{t} \\
H_{t}\tilde{\mu}_{t}
\end{pmatrix},\begin{pmatrix}
\tilde{\Sigma}_{t} & \tilde{\Sigma}_{t}H_{t}^{'} \\
H_{t}\tilde{\Sigma}_{t} & H_{t}\tilde{\Sigma}_{t}H_{t}^{'} + R_{t}
\end{pmatrix}\right)
\end{equation}

\noindent
Then the updated filtering distribution is $x_{t}|y_{1{:}t} \sim \mathcal{N}(\hat{\mu}_{t}, \hat{\Sigma}_{t})$ where  $\hat{\mu}_{t}$ and $\hat{\Sigma}_{t}$ are given by

\begin{align}\label{eq:6}
\hat{\mu_t} := \tilde{\mu_t} + K_{t}(y_{t} - H_{t}\tilde{\mu_t}) ,\nonumber\\
\hat{\Sigma_t} := (I_{n} - K_{t}H_{t})\tilde{\Sigma}_{t},
\end{align}

\noindent
with $K_t := \tilde{\Sigma}_{t}H_{t}^{'}(H_{t}\tilde{\Sigma_t}H_{t}^{'} +R_{t})^{-1}$ being the Kalman Gain Matrix. For large $n$ and $m_t$, computing the matrices in \eqref{eq:6} is computationally expensive and often leads to numeric instability. 

%It should be noted that the dimensionality of $\tilde{\Sigma_t}$, $\hat{\Sigma_t}$ is $n \times n$, and the dimensionality of the inverse term in the Kalman gain matrix is $m_{t} \times m_t$. For large $n$ and $m_t$, computing these matrices is extremely expensive.

\subsection{Ensemble Kalman Filter}\label{sec:enkf}

The idea of EnKF is to take an ensemble of size $N$ from the filtering distribution at $t-1$. This ensemble is denoted as $\hat{x}^{(0)}_{t-1}, \ \ \hat{x}^{(1)}_{t-1}, ..., \hat{x}^{(N)}_{t-1} \sim \mathcal{N}_{n}\left(\hat{\mu}_{t-1}, \hat{\Sigma}_{t-1}\right)$. In the forecast step of EnKF, \eqref{eq:2} is applied to the ensemble members to obtain their evolution from $t-1$ to $t$. That is 
\begin{equation}\label{eq:7}
\tilde{x}^{(i)}_{t} = M_{t}\hat{x}_{t-1}^{(i)} + \eta_{t}^{(i)}, \ \ \eta_{t}^{(i)} \sim \mathcal{N}(0, Q_t), \ \ \ \ i = 1, \ldots, N
\end{equation}

It can also be shown that $\tilde{x}^{(i)}_{t} \sim \mathcal{N}(\tilde{\mu}_{t}, \tilde{\Sigma}_{t})$. Similar to the update step of the Kalman Filter, all the members of this ensemble must be updated when the measurement at time step t becomes available. To update these ensemble members, first, a sample is obtained for the measurement error sequences that is $\epsilon_{t}^{(1)}, \epsilon_{t}^{(2)}, ..., \epsilon_{t}^{(N)} \sim \mathcal{N}_{m_t}(0, R_t)$. Then using these simulated measurement errors $N$ perturbed observations $\tilde{y_{t}}^{(1)}, \tilde{y_{t}}^{(2)}, \ldots, \tilde{y_{t}}^{(N)}$ are obtained using $\tilde{y_{t}}^{(i)} = H_{t}\tilde{x}^{(i)}_{t} + \epsilon_{t}^{(i)}$. Since the joint distribution of  ($\tilde{x}^{(i)}_{t}$, $\tilde{y_{t}}^{(i)}$) is the same as in \eqref{eq:5}, the updating equations are  obtained by shifting the forecasted ensemble in \eqref{eq:7} as follows

\begin{equation}\label{eq:8}
\hat{x}^{(i)}_{t} = \tilde{x}_{t}^{(i)} + K_{t}(y_t - \tilde{y}^{(i)}_{t}), \ \ \ \ i = 1, \ldots, N
\end{equation}

It can be easily shown that $\hat{x}^{(i)}_{t} \sim \mathcal{N}_{n}(\hat{\mu}_{t}, \hat{\Sigma}_{t})$. %Recall from the previous section that the Kalman Gain matrix, required in \eqref{eq:8} to compute the filtering distribution, contains the forecast covariance matrix at time $t$, $\tilde{\Sigma_t}$ term which is $n\times n$ dimension matrix. So instead of calculating this huge matrix, the sample covariance matrix of the forecasted ensemble from \eqref{eq:7} is used to estimate $\tilde{\Sigma}_{t}$. 
The computational benefit comes from the fact that instead of computing the Kalman gain matrix in \eqref{eq:8} explicitly, the sample covariance matrix of the forecasted ensemble ($\tilde{S_t}$, say) is used to estimate the Kalman Gain matrix as $\hat{K_t} := \tilde{S}_{t}H_{t}^{'}(H_{t}\tilde{S_t}H_{t}^{'} +R_{t})^{-1}$.

\subsection{KF and EnKF for Deep Learners}\label{sec:kfindl}

Although conventional KF is only suitable for estimating parameters in linear state-space models, several extensions have been proposed to generalize KF in nonlinear settings. For instance, \cite{rivals1998recursive} used extended KF to train feed-forward NN. \cite{wan2000unscented} introduced the unscented KF that better approximates nonlinear systems while making it amenable to the KF framework. \cite{anderson2001ensemble} used the concept of state augmentation that offered a generic method to handle nonlinearity in state-space models via the KF framework. \cite{iglesias2013ensemble} utilized this state augmentation technique to develop a generic method to train ANNs. They derived the state-augmented KF's forecast and updated equations in ANNs, thereby providing the algebraic framework to train DLs using the Ensemble Kalman Filters approach. %The general matrix state-space equations provided by \cite{choukroun2006kalman} provide a way to represent the ensemble learning problem mathematically. Using properties of the `vec` operator and Knonecker product, they first boil down the matrix state-space model into its vector analog, which can then be solved using standard vector form Kalman Filter update equations.
%The use of Kalman Filters and their variants, such as the Ensemble Kalman Filters, have been surprisingly sparse in the deep learning literature. 
These equations were subsequently used by \cite{yegenoglu2020ensemble} to train a Convolutional Neural Network using EnKF. Furthermore, \cite{chen2019approximate} %train a single hidden layer neural network for a regression task using the method outlined in
also used the updating equations in \cite{iglesias2013ensemble} to train a single hidden layer ANN and demonstrated how using state augmentation one can estimate the measurement error variance. State-augmented EnKF formulation also estimated parameters in LSTMs \citep{chen2018approximate}.% also demonstrate the use of Ensemble Kalman Filters in training Long Short Term Memory (LSTM) models, which are deep neural network architectures developed to solve natural language processing tasks such as tweet classification and sentiment analysis. 

All the foregoing models offer techniques to estimate parameters of complex nonlinear DLs using the EnKF framework. However, they are unsuitable when we have multiple feature representations. We want to approximate a DL with a multi-arm ANN trained via EnKF, as discussed in section \ref{sec:datadesc}.

\section{Methodology}\label{sec:method}
First, we offer a generic construction of the proposed MEnKF-ANN procedure and describe how this method could be deployed to solve the problem in section \ref{sec:datadesc}. We will use the following notations. $Y\in \mathcal{R}$ is our target response. We have a total of $m=\sum_{t=1}^T m_t $ training instances, with $m_t$ being the number of training data points in the $t^{th}$ batch. $v_t^f \in \mathcal{R}^p$ and $v_t^g\in \mathcal{R}^q$ denote two different representations of the features (possibly of different dimensions) for the $t^{th}$ batch of data. Consider two ANNs, denoted by $f$ and $g$. Assume that the neural networks f and g have $n_f$, $n_g$ number of learnable parameters. For illustrative purposes, we will assume $n_f=n_g$. If $n_f\neq n_g$, we can use suitable padding when updating the weights. In the $t^{th}$ batch of data, we assign the feature sets $v_t^f$ and $v_t^g$ to networks $f$ and $g$, respectively. We denote $w_t^{f}$ and $w_t^{g}$ to be the updated weights for the neural network $f$ and $g$, respectively.%, where $w_t^{f}$ has been padded with zeros to make it the same dimension as $w_t^{g}$.

\subsection{Matrix Kalman Filter based Multi-arm ANN}\label{9}
Consider the state matrix, $X_t$, associated with the $t^{th}$ batch of data  given by

\begin{equation}\label{eq:20}
X_t^{(m_t + n_g + 1) \times 2} =
\begin{bmatrix}
f(v_t^f, w_t^{f}),  \ \ g(v_t^g, w_t^{g}) \\
w_t^{f}, \ \ w_t^{g} \\
0, \ \ a_t
\end{bmatrix}
\end{equation}

\noindent
where $a_t$ and $b_t$ are real-valued scalar parameters. Define $H_t^{m_t \times (m_t + n_g + 1)} = [I_{m_t}, 0_{m_t \times (n_g + 1)}]$ and $G_t^{2\times1} = [1-\sigma(a_t), \ \ \sigma(a_t)]^T$ where $\sigma(.):\mathcal{R}\rightarrow{[0,1]}$, with the sigmoid function being a popular choice of $\sigma(.)$. Additionally, define $\Theta_{t-1} = I_{m_t + n_g +1}$ and $\psi_{t-1} = I_2$. We are now in a position to define the Matrix Kalman Filter.
The measurement equation is given by: 
\begin{equation}\label{eq:14}
Y_t = H_{t}X_{t}G_{t} + \epsilon_{t}
\end{equation}
with the state evolution equation being 
\begin{equation}\label{eq:13}
X_t = \Theta_{t-1}X_{t-1}\psi_{t-1} + \eta_{t}
\end{equation}

\noindent
Writing in $vec$ format,  \eqref{eq:13} becomes
\begin{equation}\label{eq:16}
x_t=vec(X_t) = (\psi^{T}_{t-1} \otimes \Theta_{t-1}) vec(X_{t-1}) + vec(\eta_{t})
\end{equation}

\noindent
 Now letting $\phi_{t-1} = \psi^{T}_{t-1}\otimes \Theta_{t-1}$ and $\tilde{\eta}_t=vec(\eta_t)$ we get from \eqref{eq:16}
\begin{equation} \label{eq:17}
x_{t} =  \phi_{t-1}x_{t-1} + \tilde{\eta}_{t}
\end{equation}

\noindent
\eqref{eq:14} can similarly be compactified as 
\begin{equation}\label{eq:18}
y_t = \mathcal{H}_{t} x_{t} + \epsilon_{t}
\end{equation}

\noindent
where $ \mathcal{H}_{t}= G^{T}_{t}\otimes H_{t}$. Observe that\eqref{eq:18} and \eqref{eq:17} have the same form as the standard representation of linear state space model described in \eqref{eq:1} and \eqref{eq:2}. Therefore, we can get the matrix state space model's solution by converting it to the vector state space model and then using EnKF to approximate the updating equations. We direct the audience to \citep{choukroun2006kalman} for more details on Matrix Kalman Filters.

\subsection{Interpreting MEnKF-ANN and a Reparametrization}

The above construction of $X_t$, $H_t$, and $G_t$ performs automatic model averaging while training. First, consider the matrix multiplication of $H_{t}X_{t}$ from \eqref{eq:14}. This would be a $m_t \times 2$ dimensional matrix in which the first column is the prediction, for the $t^{th}$ batch, from the neural network $f$ and the second column is the prediction from the neural network $g$. Post multiplication by $G_t$ would take the weighted average of each row in $H_tX_t$ where the weights are defined inside the $G_t$ matrix. %Therefore, each row of this matrix would have the predictions from neural network $f$ and $g$ for the corresponding sample from $t^{th}$ batch $v_t$. 
Now consider the matrix multiplication of $H_{t}X_{t}G_{t}$ from \eqref{eq:14}

\begin{eqnarray}
H_{t}X_{t}G_{t} &=& 
\begin{bmatrix}
f(v_t^f, w_t^{f}),  \ \ g(v_t^g, w_t^{g})
\end{bmatrix}
\begin{bmatrix}
1-\sigma(a_t) \\
\sigma(a_t)
\end{bmatrix}\nonumber\\
&=&
\begin{bmatrix}
(1-\sigma(a_t))  f(v_t^f, w_t^{f}) + \sigma(a_t)  g(v_t^g, w_t^{g})
\end{bmatrix}
\label{eq:22}
\end{eqnarray}

%\begin{equation}\label{eq:23}
%H_{t}X_{t}G_{t} = 
%\begin{bmatrix}
%(1-\sigma(a_t))  f(v_t, w_t^{f}) + \sigma(a_t)  g(v_t, w_t^{g})
%\end{bmatrix}
%\end{equation}

\noindent
\eqref{eq:22} clearly demonstrates how our construction explicitly performs model averaging across the batches with $1-\sigma(a_t)$ and $\sigma(a_t)$ being the convex weights allocated to the ANNs $f$ and $g$, respectively. 

Although the foregoing construction connects Matrix KF formulation with multi-arm ANN and performs explicit model averaging, it suffers from a computational bottleneck. Using \eqref{eq:17} and \eqref{eq:18}  the estimated Kalman Gain Matrix would be $K_{t} = \tilde{S}_{t}\mathcal{H}_{t}^{T}(\mathcal{H}_t\tilde{S}_{t}\mathcal{H}_{t}^{T} + \sigma^{2}_{y}I_{m_t})^{-1}$ . %need not be computed for each member of the ensemble in EnKF but rather computed just once and then used for updating each of the ensemble members.% 
However, in the above parameterization we have  $G_t = [1-\sigma(a_t), \ \ \sigma(a_t)]^T$ and %$\mathcal{H}_{t}$ is a function of $G_t$%
$ \mathcal{H}_{t}= G^{T}_{t}\otimes H_{t}$. This would require computation of the estimated Kalman Gain matrix for each member in EnKF since, at any given iteration of our MEnKF-ANN, we have an $a_t$ for each member of the ensemble. Thus computation complexity associated with the Kalman Gain computation increases linearly with the size of the ensemble in the above parametrization of the MEnKF-ANN. %if we have an ensemble size of $N$, we will have $N$  $G_t$s, which will require computation of the matrix products for the Kalman Gain Computation $N$ times. Evidently, computation complexity increases at least linearly \hl{(Since there are two matrixes, would it now be quadratically)} with the size of ensemble the above parametrization of the MEnKF-ANN.

To alleviate this computational bottleneck, consider the following parametrization:
\begin{align}\label{eq:24}
X_t =
\begin{bmatrix}
(1-\sigma(a_t))  f(v_t^f, w_t^{f}),  \ \sigma(a_t)  g(v_t^g, w_t^{g}) \\
w_t^{f}, \ \ w_t^{g} \\
0, \ \ a_t
\end{bmatrix}
\end{align}
\\
and $G_t = [1, \ \ 1]^T$. We still have explicit model averaging in the measurement equation, i.e.,
\begin{align}\label{eq:26}
H_{t}X_{t}G_{t} = 
\begin{bmatrix}
(1-\sigma(a_t)) f(v_t^f, w_t^{f}) + \sigma(a_t)  g(v_t^g, w_t^{g})
\end{bmatrix}
\end{align}
but $\mathcal{H}_t$ does not depend on $a_t$. Therefore the matrix products for the Kalman Gain computation can now be computed once for each batch. %and then used to update each ensemble member. 

%\subsection{MEnKF-ANN that can learn $\sigma^{2}_{y}$}
Turning to the variance parameter in the measurement equation \eqref{eq:18}. Assume $\epsilon_{t} \sim \mathcal{N}_{m_{t}}(0,\nu^{2}_{y}I_{m_t} )$
To estimate $\nu^{2}_{y}$, we augment the state vector  as follows: 

\begin{align}\label{eq:27}
X_t^{(m_t + n_g + 2) \times 2} =
\begin{bmatrix}
(1-\sigma(a_t))  f(v_t^f, w_t^{f}),  \ \sigma(a_t)  g(v_t^g, w_t^{g}) \\
w_t^{f}, \ \ w_t^{g} \\
0, \ \ a_t \\ 
0, \ \ b_t
\end{bmatrix}
\end{align} 

where $\nu^{2}_{y} = \log(1 + e^{b_t})$ and $H_t$ in \eqref{eq:14} now becomes $ [I_{m_t}, 0_{m_t \times (n_g + 2)}]$. We used a softplus transformation of $\nu^2_y$, instead of the usual log transformation for computational stability.

\subsection{Connecting MEnKF-ANN with DL}
Recall that our dataset consists of string-valued gene sequences associated with experimentally determined PULs, with the response being the carbohydrate substrates utilized by the said microbe. Since we consider only two categories of PULs, we have a binary classification problem. An LSTM trained with a binary cross-entropy loss is the approximand DL in our case. Suppose $p$ is the probability of observing a sample of a particular category. In that case, the trained  LSTM produces $\hat{p}$ for each training instance, along with an embedding of the associated gene sequences. Our MEnKF-ANN approximator uses $logit(\hat{p})$ as the target response. The LSTM embedding of the gene sequences is fed into one arm of the approximator while the other arm ingests Doc2Vec encoding of the gene sequences. Thus, our MEnKF-ANN approximates the probabilities estimated by an LSTM. The convex weights $\sigma(a)$ ascertain which embedding has more predictive power. Clearly, MEnKF-ANN operates as a model stacker, and the predictive uncertainty interval that it produces, by default, around its target approximand quantifies how well simpler ANNs, fitted without backpropagation, can approximate a deep learner.

To initialize the ensemble in the approximator, we draw the members in the state vector \eqref{eq:27} from $\mathcal{N}_{2(m_t+n_g+2)}(\mathbf{0}, \nu^2_{x}I)$, where $\nu^2_x$ is a tuning parameter that plays a key role in controlling the spread of the ensemble members and the dimension of $I$ matches with the dimension of normal distribution. Following \cite{chen2018approximate, chen2019approximate}, we assume the state transition is deterministic, i.e., $x_t=\phi_{t-1}x_{t-1}$ and hence we do not have the variance parameter corresponding to $\tilde{\eta}$ in the augmented state vector. When we reach the $t^{th}$ batch of data, for the $i^{th}$ member in the ensemble ($i=1,2,...,N$), we update each element in the augmented state vector $w^{f,(i)}_{t}$, $w^{g,(i)}_{t}$, $a^{(i)}_{t}$, $b^{(i)}_{t}$  using the updating equation \eqref{eq:8} suitably modified to handle deterministic state transition. %$t^{th}$ batch of data.
%$\sigma^2_{w}$ is the prior variance for the ensembles that are drawn at the time step zero from $\mathcal{N}_{d}(\mathbf{0}, \sigma^2_{w}I_d)$. $\sigma^{2}_{w_{vec}}$ is the variance for the error terms $\tilde{\eta}_t$ from \eqref{eq:17} that are needed in the computation of the forecast step of the algorithm and are drawn from $\mathcal{N}_{(2\times(m_t + n_g + 2))}\left(0, \sigma^{2}_{w_{vec}}I_{(2\times(m_t + n_g + 2))}\right)$. $\sigma^{2}_{y}$ is the variance for the error terms $\epsilon_{t}$ from \eqref{eq:18} that are required in the computation of the update step of the algorithm and are drawn from $\mathcal{N}_{m_t}(0, \sigma^{2}_{y}I_{m_t})$. $w^{(i)}_{f,t}$, $w^{(i)}_{g,t}$, $a^{(i)}_{t}$, $b^{(i)}_{t}$ represent the updated weights using the $t^{th}$ batch from the $i^{th}$ ensemble member for the neural networks $f$, $g$, the averaging weight that defines the $G_t$ matrix, and the parameter for estimating $\sigma^2_{y}$.--- \textbf{this needs to be checked}

\section{Simulations}\label{sec:simulation}
We conducted extensive simulations to assess how well our MEnKF-ANN can approximate an LSTM binary classifier. This simulation exercise aims to demonstrate that our MEnKF-ANN is not only ``adequate'' in approximating the probabilities produced by LSTM but can also capture the ``true'' probabilities that generate binary labels. We compute the coverage and width of the prediction intervals of the target probabilities in the test set to assess the ``adequacy'' of the approximator. Then, we compare this coverage and width with those computed directly via an LSTM  trained with MC dropout. Admittedly, the prediction intervals obtained from the latter are different from those computed from MEnKF-ANN. However, if the ground truth probabilities are known, an adequate approximator should be able to achieve near-nominal coverage when the approximand is not misspecified.

Our simulation strategy mimics the focal dataset and uses the gene sequences associated with the original PULs to generate labels. As mentioned above, we extracted $\hat{p}$ from the LSTM trained on the original dbCAN-PUL data. We call this LSTM the \textit{true LSTM}. We consider $\hat{p}$ the true probabilities for synthetic data generation. We then use noisy copies of $\hat{p}$ to generate a synthetic label in the following way: generate $logit(\tilde{p}_i^{(j)})=logit(\hat{p}_i)+\epsilon_i^{*(j)},\; i=1,2,...,m, j=1,2,...,J$, where  $J$ is the number of the simulated dataset and $m$, is the number of data points in each simulated set, the perturbation $\epsilon_i^{*(j)}$ are iid Normal(0,$0.01^{2}$). We generate synthetic labels $\tilde{Y}$ by thresholding $\tilde{p}_i^{(j)}$ at 0.5, i.e $\tilde{Y}_i^{(j)}=I(\tilde{p}_i^{(j)}>0.5)$. Then the simulated dataset consists of $D^{(j)}=\{\bm{F},\tilde{Y}^{(j)}, \; j=1,2,...,J\}$, where $\bm{F}$ is the set of original gene sequences from dbCAN-PUL.

\begin{figure}[!t]%
\centering
\includegraphics[width =  0.7\textwidth]{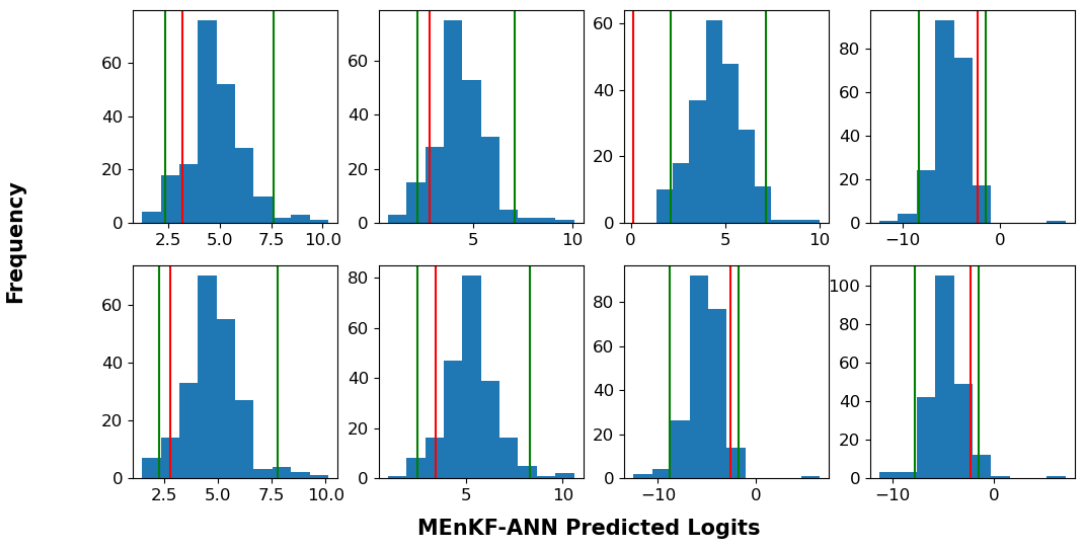}
\caption{True Logits Superimposed on Predicted Logits from MEnKF-ANN using LSTM and Doc2Vec embedding}\label{fig1}
\label{fig:boxplots}
\end{figure}

Now in each $D^{(j)}$, we train a second LSTM (with two dropout layers) and extract $\tilde{\tilde{p}}_i^{(j)}, i=1,2,...,m$ and the embedding of the gene sequences. We call these LSTMs, trained on $D^{(j)}$, the \textit{fitted LSTMs}. Note that the embeddings from \textit{fitted LSTMs} could potentially be different from those obtained from the \textit{true LSTM}. We denote the embedding from \textit{fitted LSTMs}  by $v_{i}^{(j),f},\; j=1,2,...J$. Our MEnKF-ANN is constructed to approximate the \textit{fitted LSTMs}. To that end, the approximator uses $logit(\tilde{\tilde{p}}_i^{(j)})$ as the target response. $v_i^{(j),f}$ are supplied as features to one arm of the ANN, the other arm ingests $v_i^{(j),g}$ - the Doc2Vec embedding of $\bm{F}$. Once the MEnKF-ANN is trained, we use a hold-out set in each simulated dataset to generate predictive probabilities from the forecast distribution for each member in the KF ensemble and compute the empirical 95\% predictive interval at $logit^{-1}$ scale. To measure the adequacy of MEnKF-ANN, we compute the proportion of times the foregoing predictive interval contains $\hat{p}$ in held-out test data. We expect this coverage to be close to the nominal 95\%, and the average width of these intervals should not be greater than 0.5. Additionally, observe that the data-generating model uses LSTM embedding of $F$; hence, using Doc2Vec embedding as input is misspecification. Consequently, we expect the average model weight associated with $v^f$ to be larger than $v^g$. Table \ref{tab1}  shows the performance of MEnKF-ANN in terms of coverage, the average width of prediction intervals, and average LSTM weight under two specifications of ensemble size ($N$) and initial ensemble variance $(\nu^2_x)$. To compare these results, we offer the coverage and average width of the prediction intervals when both the dropout layers are activated in the \textit{fitted LSTM} during the prediction phase in  Table \ref{tab2}. Observe how MEnKF-ANN recovered the \textit{true probabilities} even better than the correctly specified LSTM with dropout. The average interval widths obtained from MEnKF-ANN are also lower than those from the \textit{fitted LSTM}. These demonstrate the adequacy of MEnKF-ANN in approximating the target DL. Additionally, we observe that the average LSTM model weight is $\approx 1$ indicating the ability of our approximator to identify the correctly specified data-generating model. Figure \ref{fig1} shows the histogram of the predictive samples obtained from the ensemble members for eight test samples in a randomly chosen replicate. The red vertical line denotes the true logits, and the green vertical lines show the fences of the 95\% prediction interval.    

Now, to demonstrate a situation where MEnKF-ANN is ``inadequate'', we supply the approximator with a completely different feature set representation. Instead of using the LSTM embedding $v^f$, we use word2vec embedding of each gene in the predictor string and take the arithmetic average of these word2vec embeddings to represent the entire sequence. We denote this feature set by $\tilde{v}^f$ and then train the MEnKF-ANN using $\tilde{v}^f$ and $v^g$ as the features and $logit(\tilde{\tilde{p}}^{(j)})$ as the target response. Evidently, MEnKF-ANN is highly misspecified. Table \ref{tab3} reports the coverage and average width of the prediction interval obtained from this model. Observing the huge width of the intervals essentially invalidates the point prediction. Such large width indicates that MEnKF-ANN may not approximate the target DL. Therefore, we caution against using the coverage and width metrics to assess the ``adequacy'' of the \textit{fitted LSTM} itself.   

\begin{table}[!t]
\caption{Performance of MEnKF-ANN using LSTM embedding and Doc2Vec  \label{tab1}}%
\begin{tabular*}{\columnwidth}{@{\extracolsep\fill}lllll@{\extracolsep\fill}}
\hline
%Dropout & 
$N$ & $\nu^{2}_{x}$ & Coverage  & Width &  LSTM weight\\
\hline
%Heavy & 
216    & 16    & 90.25$\%$  & 0.33 & 0.9997   \\
%Heavy & 
216    & 32    & 89.25$\%$  & 0.32 & 0.9999   \\
%Low & 216    & 16    & 88.00$\%$  & 42.62$\%$ & 0.9996   \\
%Low & 216    & 32    & 88.00$\%$  & 38.52$\%$ & 0.9999   \\
\hline
\end{tabular*}
\end{table}

\begin{table}[!t]
\caption{Coverage and width of prediction intervals obtained from \textit{fitted LSTM} with two dropout layers \label{tab2}}%
\begin{tabular*}{\columnwidth}{@{\extracolsep\fill}llll@{\extracolsep\fill}}
\hline
%Dropout & 
Rate & Reps & Coverage  & Width \\ %&  LSTM weight\\
\hline
%Heavy & 
0.5    & 50    & 81.25$\%$  & 0.53\\% & 0.6031   \\
%Heavy &
% 0.5    & 100    & 84.25$\%$  & 0.55\\% & 0.6228   \\
%Heavy &
0.5    & 200    & 84.50$\%$  & 0.56 \\%& 0.6362   \\
%Low & 0.5    & 50    & 86.75$\%$  & 32.49\\%\%$ & 0.6031   \\
%Low & 0.5    & 100    & 86.75$\%$  & 31.09\\%\%$ & 0.6228   \\
%Low & 0.5    & 200    & 86.75$\%$  & 33.10\\%\%$ & 0.6362   \\
\hline
\end{tabular*}
% \begin{tablenotes}%
% \item Source: The average accuracy and the standard error are calculated using the test folds in a ten-fold cross validation.
% \item[$^{1}$] This corresponds to the EnKF-Stacker model with the hyperparameter settings of $0.01$ for the target variance and $1$ for dimension reduction factor. It has 668 number of learnable parameters between the two constituent feed forward neural networks and 1 learnable parameter for the meta-model which averages the predictions from the two constituent models. 
% \item[$^{2}$] This corresponds to the Stacking model trained using backpropagation where the constituent models are the same as in the EnKF-Stacker model. It has 668 number of learnable parameters between the two constituent feed forward neural networks and 3 learnable parameters for the meta-model. 
% \end{tablenotes}
\end{table}

\begin{table}[!t]
\caption{Performance of MEnKF-ANN using Word2Vec and Doc2Vec  \label{tab3}}%
\begin{tabular*}{\columnwidth}{@{\extracolsep\fill}lllll@{\extracolsep\fill}}
\hline
%Dropout & 
$N$ & $\nu^{2}_{x}$ & Coverage  & Width &  Word2Vec weight\\
\hline
%Heavy & 
216    & 16    & 96.25$\%$  & 0.83 & 0.9155  \\
%Heavy & 
216    & 32    & 94.25$\%$  & 0.84 & 0.9787   \\
%Low & 216    & 16    & 96.50$\%$  & 84.34$\%$ & 0.9377  \\
%Low & 216    & 32    & 95.25$\%$  & 85.63$\%$ & 0.9755   \\
\hline
\end{tabular*}
% \begin{tablenotes}%
% \item Source: The average accuracy and the standard error are calculated using the test folds in a ten-fold cross validation.
% \item[$^{1}$] This corresponds to the EnKF-Stacker model with the hyperparameter settings of $0.01$ for the target variance and $1$ for dimension reduction factor. It has 668 number of learnable parameters between the two constituent feed forward neural networks and 1 learnable parameter for the meta-model which averages the predictions from the two constituent models. 
% \item[$^{2}$] This corresponds to the Stacking model trained using backpropagation where the constituent models are the same as in the EnKF-Stacker model. It has 668 number of learnable parameters between the two constituent feed forward neural networks and 3 learnable parameters for the meta-model. 
% \end{tablenotes}
\end{table}

\section{Application}\label{sec:application}
%We extract the dataset from the dbCAN-PUL database \citep{ausland2021dbcan} that contains experimentally verified PULs and the corresponding GenBank sequences of these PULs along with known target carbohydrate substrates \hl{is this correct?}. As mentioned earlier, data on such experimentally derived PUL along with their target  substrate is  limited, we have a total of \textbf{~600} data points. Figure \textbf{zz} below shows the frequency distribution of various target substrate in the dataset. Clearly, we do not have sufficient sample to train a complex DL to classify all the target substrate. Hence we propose to classify only the top 2 most-frequently occurring target substrate- Xylan and Pectin and train a binary classifier. 
Recall that our focal dataset consists of $n=74$ samples belonging to Xylan and Pectin. However, training an LSTM on a small sample size would require aggressive regularization, even with this reduced label space. Therefore, we draw on an extensive collection of unlabelled data containing gene sequences associated with CAZyme gene clusters (CGC) computationally predicted from genomic data \citep{huang2018dbcan, zheng2023dbcan}. Although this unlabelled data contain approximately 250K CGC gene sequences, unlike experimentally characterized PULs, these sequences do not have known carbohydrate substrate information. They hence cannot be directly used for classification purposes. We, therefore, use this unlabelled dataset to learn the word2vec embeddings of each gene appearing in the unlabelled dataset. These embeddings are then used to initialize the embedding layer of the target LSTM classifier. 

Turning to the labeled dataset, instead of performing full cross-validation, we resort to subsampling procedure \citep{politis1999subsampling}. We take a subsample of sixty-six instances for training and hold eight instances for testing purposes. The subsample size ($b$) is chosen such that $b(n)/n \approx 8\sqrt{n}/n \rightarrow 0,\;$ as $n\rightarrow \infty$. Although the subsampling theory requires generating $\begin{pmatrix}     n\\b \end{pmatrix}$ replicates, the computational cost for generating $\approx 10^{11}$ replicates, in our case, is prohibitive. Instead, we generate 50 independently subsampled replicates comprising training and testing sets of sizes 66 and 8, respectively. In each replication,  an LSTM with the same architecture is trained on the foregoing 66 training instances. Under this scheme, the probability that the $i^{th}$ instance in our dataset appears at least once in the test set is $\approx 99.6\%$.   %of  to generate 50 replicates of the labeled dataset and train LSTM classifier using a randomly selected subsample of 66 data points. The remaining 8 data points, in each replicate, are held out for testing purpose. 

\begin{figure}[!t]%
\centering
\includegraphics[width =  0.7\textwidth]{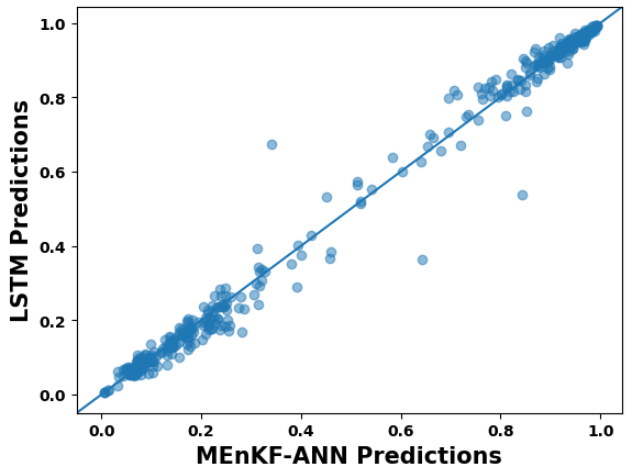}
\caption{Scatterplot of First LSTM Predicted Probabilities vs. EnKF Predicted Probabilities}\label{fig2}
\label{fig:enkfvslstmpredictions}
\end{figure}

\begin{figure}[!t]%
\centering
\includegraphics[width =  0.7\textwidth]{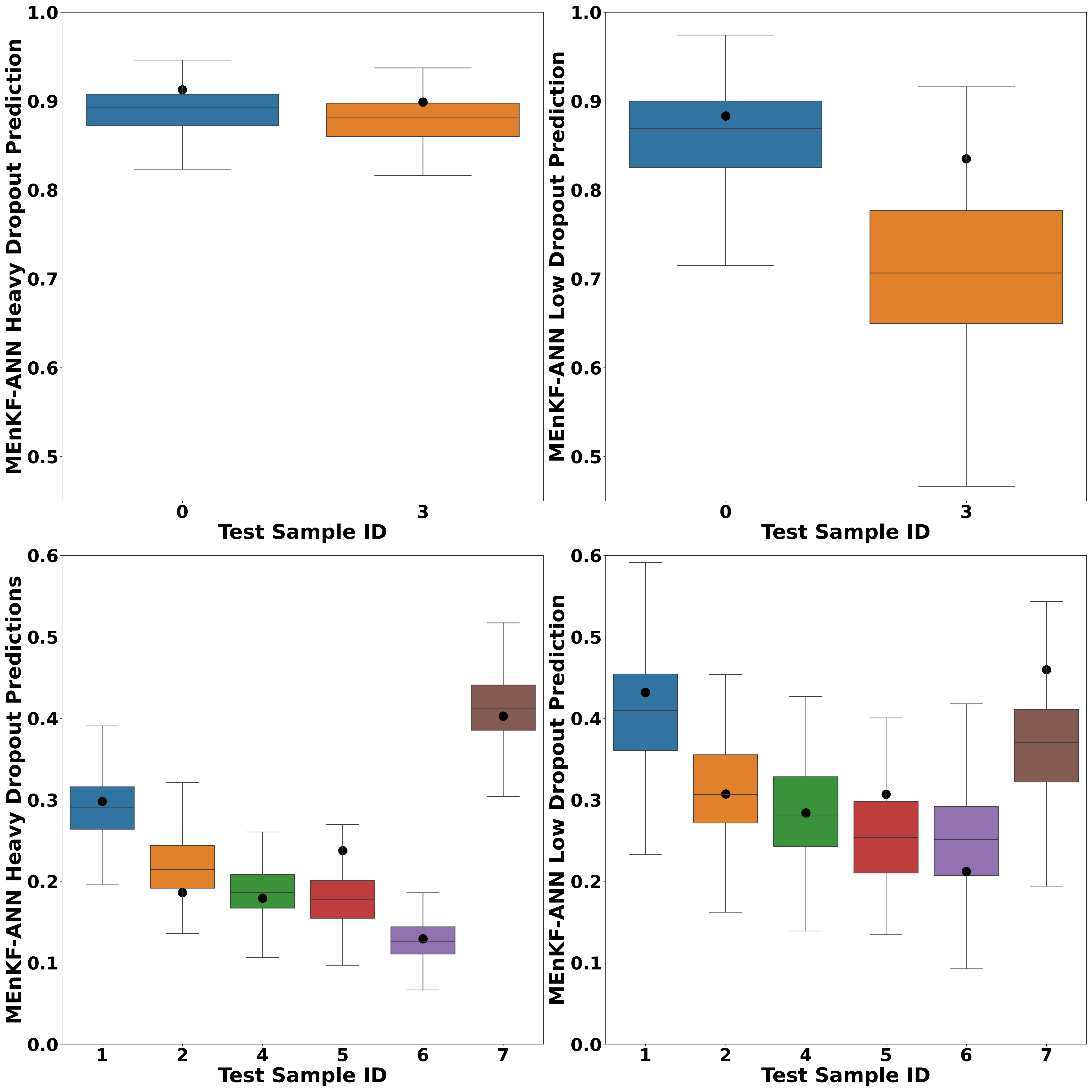}
\caption{Boxplots showing the MEnKF-ANN Predictions superimposed with the ground truth value for Heavy and Low Dropout}%\label{fig1}
\label{fig:menkfboxplots}
\end{figure}

The LSTM-estimated probabilities of observing a \textit{Pectin} substrate are extracted from each replicate. These probabilities are logit transformed and used as the target response for our MEnKF-ANN approximator. We feed the LSTM embedding and Doc2Vec embedding of the gene sequences into the two arms of the approximator along with the foregoing logit-transformed estimated probabilities. We then generate predictions on the held-out test data points in each replicate. Finally, we compare the LSTM-prediction of probabilities with those generated by  MEnKF-ANN generated predictions. The average MAE and the proportion of times a 95\% prediction interval contains the LSTM-generated predictions in the held-out data set, under two different MEnKF-ANN hyperparameter choices are shown in Table \ref{tab4} indicating that our approximator can be adequately used to generate the predictions. We do not report the LSTM weights estimated by MEnKF-ANN because, as we observed in the simulation (Table \ref{tab1}), the approximator overwhelmingly prefers the LSTM embeddings.   Figure \ref{fig:enkfvslstmpredictions} shows the scatter plot of MEnKF-ANN-predicted and LSTM-predicted probabilities for the held-out data across 50 replicates. Figure \ref{fig:menkfboxplots} shows the boxplots associated with MEnKF-ANN predictions for the same set of test samples for which LSTM generated prediction boxplots were shown in the left column of Figure \ref{fig:boxplots2}. Evidently, MEnKF-ANN can adequately approximate the target LSTM.

Turning to the stability of prediction intervals, Table \ref{tab5} shows the average width of the 95\% prediction intervals obtained under two configurations of LSTM and their respective MEnKF-ANN approximators. LSTM$_1$ has two dropout layers (one in the LSTM layer and one before the final prediction layer) with a 50\% dropout rate and 200 replicates. LSTM$_2$ has one dropout layer (in the LSTM layer) with a 50\% dropout rate and 200 replicates. MEnKF-ANN$_{11}$ approximates LSTM$_1$ with 216 ensemble members and $\nu^2_x=16$, MEnKF-ANN$_{12}$ also approximates LSTM$_1$, but now with 216 ensemble members and $\nu^2_x=32$. Similarly, MEnKF-ANN$_{21}$ and MEnKF-ANN$_{22}$ approximates LSTM$_2$ with 216 ensemble members and $\nu^2_x=16$ and $\nu^2_x=32$, respectively. Observe that the variation in the average width between LSTM$_1$ and LSTM$_2$ is considerably higher than the variation between MEnKF-ANN$_{11}$ and MEnKF-ANN$_{21}$ or between MEnKF-ANN$_{12}$ and MEnKF-ANN$_{22}$. This indicates that the approximator produces more stable prediction intervals than obtaining prediction by activating the dropout layer during prediction.  %their the MEnKF-ANN generated prediction interval compares favorably with the average the width of the prediction interval generated using LSTM with MC-dropout. Table \textbf{llll} below shows a comparison of this average width under various hyper-parameter setting. Observe that the computational time for MEnKF-ANN is only a fraction of that required by droup-out method. Evidently, MEnKF-ANN is a fast and adequate approximator of the focal ANN.

Finally, we demonstrate how MEnKF-ANN can naturally handle two predictive models with potentially different feature sets. This situation is relevant because, owing to the small sample size, we can train a shallow learner (ANN with backpropagation, for instance) that takes Doc2Vec representation of gene sequences as predictors to estimate the probabilities of observing the \textit{Pectin} substrate. Now, we can average the probabilities estimated by the LSTM ($\hat{p}_{LSTM}$, say) and ANN ($\hat{p}_{ANN}$, say) to produce a model-averaged estimated probability of observing \textit{Pectin} ($\hat{\bar{p}}$, say). However, how would we attach uncertainty to $\hat{\bar{p}}$ ? The multi-arm construction of MEnKF-ANN provides a natural solution in this situation. We supply, as described in the foregoing sections, LSTM embeddings and Doc2Vec embeddings to the two arms of MEnKF-ANN but use $logit(\hat{\bar{p}})$ as the target response here. Thus MEnKF-ANN is now approximating the average output of two primary models. These primary models are trained on the same response variable but use two different representations of features. Table \ref{tab6} shows the performance of MEnKF-ANN in this situation for some combinations of $N$ and $\nu^2_x$. The coverage is measured with respect to $\hat{\bar{p}}$ on the test sets. Although the average width and MAE are larger than those reported in Table \ref{tab4}, we observe that the LSTM weights $\approx$ 0.5, which is what we would expect because MEnKF-ANN is \textit{seeing} equally weighted outputs from LSTM and ANN. 

\begin{table}[!t]
\caption{Performance of MEnKF-ANN using LSTM embedding and Doc2Vec for dbCAN-PUL data  \label{tab4}}%
\begin{tabular*}{\columnwidth}{@{\extracolsep\fill}llllll@{\extracolsep\fill}}
\hline
 $N$ & $\nu_x^{2}$ & Coverage  & Width &  MAE & CPU Time\\
\hline
216    & 16    & 90.50$\%$  & 0.1024 & 0.0200 & 2.39 mins  \\
216    & 32    & 85.50$\%$  & 0.0850 & 0.0161 & 3.67 mins \\
%Low & 216    & 16    & 92.75$\%$  & 0.1185 & 0.0198 & 2.25  \\
%Low & 216    & 32    & 93.25$\%$  & 0.1077 & 0.0162 & 3.41 \\
\hline
\end{tabular*}
\end{table}

\begin{table}[!t]
\caption{Comparison of the average width of prediction interval LSTM + MC Dropout and MEnKF-ANN approximator for each LSTM  \label{tab5}}%
\begin{tabular*}{\columnwidth}{@{\extracolsep\fill}lllll@{\extracolsep\fill}}
\hline
Target model & Average Width & Approximator & Average Width \\
\hline
LSTM$_1$  & 0.492 & MEnKF-ANN$_{11}$ & 0.102 \\
          &      & MEnKF-ANN$_{12}$ & 0.085\\
LSTM$_2$  & 0.371 & MEnKF-ANN$_{21}$& 0.119 \\       
         &        & MEnKF-ANN$_{22}$& 0.108\\
%Heavy & 0.5    & 200    & 100.00$\%$  & 0.4924 & 0.0782 & 0.4953   \\
%Low & 0.5    & 200    & 97.75$\%$  & 0.3708 & 0.0691 & 0.4873   \\
\hline
\end{tabular*}
\end{table}

\begin{table}[!t]
\caption{Performance of MEnKF-ANN trained on the averaged probability of LSTM and shallow ANN  \label{tab6}}%
\begin{tabular*}{\columnwidth}{@{\extracolsep\fill}lllllll@{\extracolsep\fill}}
\hline
 $N$ & $\nu_x^{2}$ & Coverage  & Width &  MAE & LSTM weight\\
\hline
%433    & 0.5    & 93.00$\%$  & 0.4046 & 0.0973 & 0.4358  \\
%433    & 1.0    & 90.25$\%$  & 0.3161 & 0.0771 & 0.4601  \\
433    & 2.0    & 85.50$\%$  & 0.2249 & 0.0615 & 0.4954  \\
433    & 4.0    & 78.00$\%$  & 0.1978 & 0.0684 & 0.5006  \\
%433    & 8.0    & 76.50$\%$  & 0.1987 & 0.0715 & 0.5029  \\
%Low & 216    & 16    & 92.75$\%$  & 0.1185 & 0.0198 & 2.25  \\
%Low & 216    & 32    & 93.25$\%$  & 0.1077 & 0.0162 & 3.41 \\
\hline
\end{tabular*}
\end{table}

\section{ Discussion}\label{sec:discussion}

State-augmented Kalman Filter and its variants provide a gradient-free method that can be extended to approximate popular neural network-based deep learners for regression and classification tasks. %The ensemble Kalman Filters technique could be used to speed up computation further because it does not require the computation of exact Kalman gain. 
In this article, we have developed a Matrix Ensemble Kalman Filter-based multi-arm neural network to approximate an LSTM. We have demonstrated that this technique adequately approximates the target DL regarding coverage and the average width of the prediction interval. We have also demonstrated how the in-built model averaging capability can be leveraged to attach uncertainty to the averaged predictions generated by two different models. Our simulations suggest that by using an explicit model averaging construction, our approximator can also identify its target approximand. We have also observed that the prediction intervals generated by the approximator are less sensitive to the location of dropout layers and hence provide more stable prediction intervals than obtaining predictions by activating the dropout layers within the DL itself. Admittedly, our procedure requires an additional round of training, but its fast computation time (see Table \ref{tab4}), along with its ability to emulate the approximand, adequately compensate for that. We have also deployed our approximator on a highly accessed database, dbCAN-PUL, to attach uncertainty to the predicted probabilities produced by (a) the primary LSTM model and (b) an ensemble of LSTM and ANN models. The primary LSTM and ANN models were trained to classify two carbohydrate substrates using the gene sequences characterized by the PULs of the gut microbiome. We anticipate this technique will be helpful to domain experts in assessing the reliability of predictions generated by deep learners or an ensemble of learners.

In the future, we propose to expand our model to handle more than two classes. This would enable us to utilize the information in the dbCAN-PUL database better. Another possible direction is to develop an analog of MEnKF-ANN that can directly handle binary data. Although the KF technique crucially requires Gaussianity assumption, but \cite{fasano2021closed} recently developed an extension of the KF method that can handle binary responses. We are actively investigating how this technique can be adapted to our MEnKF-ANN framework.

\section{Competing interests}
No competing interest is declared.

\section{Author contributions statement}

V.P., Y.Z., and S.G. conceived the models and experiment(s), and V.P. conducted the experiment(s). V.P. and S.G. analyzed the results. Y.Yan and Y.Yin contributed to the training data. V.P. drafted the manuscript. All authors reviewed the manuscript. Y. Yin secured the funding.
%Must include all authors, identified by initials, for example:
%S.R. and D.A. conceived the experiment(s),  S.R. conducted the experiment(s), S.R. and D.A. analysed the results.  S.R. and D.A. wrote and reviewed the manuscript.

\section{Acknowledgments}
The authors thank the anonymous reviewers for their valuable suggestions. This work is supported in part by funds from the National Institute of Health (NIH: R01GM140370, R21AI171952) and the National Science Foundation (NSF: CCF2007418, DBI-1933521). In addition, we thank the lab members for their helpful discussions. This work was partially completed utilizing the Holland Computing Center of the University of Nebraska–Lincoln, which receives support from the Nebraska Research Initiative. 

%USE THE BELOW OPTIONS IN CASE YOU NEED AUTHOR YEAR FORMAT.
\bibliographystyle{abbrvnat}
\bibliography{ref}

\end{document}